\documentclass[sigconf, anonymous=False]{acmart}

\usepackage{balance}
\def\BibTeX{{\rm B\kern-.05em{\sc i\kern-.025em b}\kern-.08emT\kern-.1667em\lower.7ex\hbox{E}\kern-.125emX}}

\usepackage{algorithm}
\usepackage{algorithmicx}
\usepackage{algpseudocode}
\copyrightyear{2019}
\acmYear{2019}
\setcopyright{acmcopyright}
\acmConference[KDD '19]{The 25th ACM SIGKDD Conference on Knowledge
Discovery and Data Mining}{August 4--8, 2019}{Anchorage, AK, USA}
\acmBooktitle{The 25th ACM SIGKDD Conference on Knowledge Discovery and Data
Mining (KDD '19), August 4--8, 2019, Anchorage, AK, USA}
\acmPrice{15.00}
\acmDOI{10.1145/3292500.3330933}
\acmISBN{978-1-4503-6201-6/19/08}

\begin{document}

%
% The "title" command has an optional parameter, allowing the author to define a "short title" to be used in page headers.
\title[Environment Reconstruction with Hidden Confounders for RL-Recommendation]{Environment Reconstruction with Hidden Confounders for Reinforcement Learning based Recommendation}
\titlenote{This work is supported by the National Key R\&D Program of China (2017YFB1002201), NSFC (61876077), Jiangsu SF (BK20170013), and Collaborative Innovation Center of Novel Software Technology and Industrialization. This work is done during the first author's internship in Didi Chuxing. Yang Yu is the corresponding author.}

%
% The "author" command and its associated commands are used to define the authors and their affiliations.
% Of note is the shared affiliation of the first two authors, and the "authornote" and "authornotemark" commands
% used to denote shared contribution to the research.
%\author{Paper ID: xxx}
%
% By default, the full list of authors will be used in the page headers. Often, this list is too long, and will overlap
% other information printed in the page headers. This command allows the author to define a more concise list
% of authors' names for this purpose.

\author{Wenjie Shang}
\affiliation{%
\institution{National Key Laboratory for Novel Software Technology\\ Nanjing University}
}
\email{shangwj@lamda.nju.edu.cn}

\author{Yang Yu}
\affiliation{%
\institution{National Key Laboratory for Novel Software Technology\\ Nanjing University}
}
   \email{yuy@nju.edu.cn}

\author{Qingyang Li}
\affiliation{%
\institution{AI Labs, Didi Chuxing}
}
\email{qingyangli@didiglobal.com}

\author{Zhiwei Qin}
\affiliation{%
\institution{AI Labs, Didi Chuxing}
}
\email{qinzhiwei@didiglobal.com}

\author{Yiping Meng}
\affiliation{%
\institution{AI Labs, Didi Chuxing}
}
\email{mengyipingkitty@didiglobal.com}

\author{Jieping Ye}
\affiliation{%
\institution{AI Labs, Didi Chuxing}
}
\email{yejieping@didiglobal.com}

%
% The abstract is a short summary of the work to be presented in the article.
\begin{abstract}
Reinforcement learning aims at searching the best policy model for decision making, and has been shown powerful for sequential recommendations. The training of the policy by reinforcement learning, however, is placed in an environment. In many real-world applications, however, the policy training in the real environment can cause an unbearable cost, due to the exploration in the environment. Environment reconstruction from the past data is thus an appealing way to release the power of reinforcement learning in these applications. The reconstruction of the environment is, basically, to extract the casual effect model from the data. However, real-world applications are often too complex to offer fully observable environment information. Therefore, quite possibly there are unobserved confounding variables lying behind the data. The hidden confounder can obstruct an effective reconstruction of the environment. In this paper, by treating the hidden confounder as a hidden policy, we propose a \emph{deconfounded multi-agent environment reconstruction} (DEMER) approach in order to learn the environment together with the hidden confounder. DEMER adopts a multi-agent generative adversarial imitation learning framework. It proposes to introduce the confounder embedded policy, and use the compatible discriminator for training the policies. We then apply DEMER in an application of driver program recommendation. We firstly use an artificial driver program recommendation environment, abstracted from the real application, to verify and analyze the effectiveness of DEMER. We then test DEMER in the real application of Didi Chuxing. Experiment results show that DEMER can effectively reconstruct the hidden confounder, and thus can build the environment better. DEMER also derives a recommendation policy with a significantly improved performance in the test phase of the real application. 
\end{abstract}



%
% Keywords. The author(s) should pick words that accurately describe the work being
% presented. Separate the keywords with commas.

%
% This command processes the author and affiliation and title information and builds
% the first part of the formatted document.
\maketitle

\section{INTRODUCTION}
In sequential recommendation problems~\cite{YeZXZGD18, YeXGD19}, where the system needs to recommend multiple items to the user while responding to the user's feedback, there are multiple decisions to be made in sequence. For example, in our application of program recommendation to taxi drivers, the system recommends a personalized driving program to each driver, and a program consists of multiple steps, where each step is recommended according to how the previous steps was followed. Therefore, recommending the program steps is a sequential decision problem, and it can be naturally solved by reinforcement learning \cite{SuttonB98}. 

\begin{figure*}[t]
	\centering
	\begin{minipage}{0.25\textwidth}
	\caption{Illustration of the graph structure and the collected data (a) in the classical environment that assumes fully observable, and (b) in the more realistic environment with an unobserved confounder.}
	\label{fig:cg_rl}
	\end{minipage}\hfill
	\begin{minipage}{0.7\textwidth}
	\includegraphics[width=\linewidth]{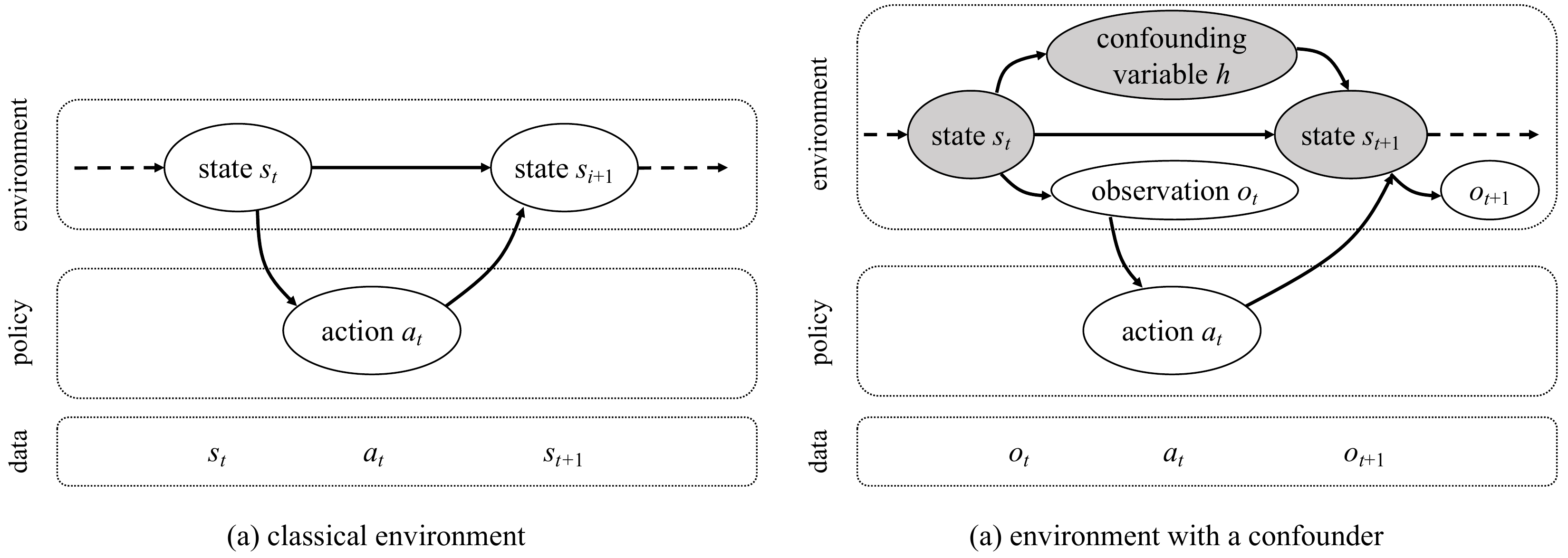}
	\end{minipage}
\end{figure*}

As a powerful tool for learning decision-making policies, reinforcement learning learns from interactions with the environment via trial-and-errors \cite{SuttonB98}. In digital worlds where interactions with the environment are feasible and cheap, it has made remarkable achievements, e.g., \cite{Mnih15,silver2016mastering}. When it comes to real-world applications, the convenience of available digital environments does not exist. It is not practical to interact with the real-world environment directly for training the policy, because of the high interaction cost and the huge amount of interactions required by the current reinforcement learning techniques. A recent study \cite{shi2018virtual} disclosed a viable option to conduct the reinforcement learning on real-world tasks, which is by reconstructing a virtual environment from the past data. As a result, the reinforcement learning process could be more efficient by interacting with the virtual environment, and the interaction cost could be avoided as well.

The environment reconstruction can be done by treating the environment as a policy that makes responses to the interactions, and employing the imitation learning \cite{schaal1999imitation, argall2009survey} to learn the environment policy from the past data, which has drawn a lot attentions recently. Comparing with using supervised learning, i.e., behavior clone, to learn the environment policy, a more promising solution in \cite{shi2018virtual}  is to formulate the environment policy learning as an interactive process between the environment and the system in it. Take the example of the commodity recommendation system, the user and the platform could be viewed as two agents interacting with each other, where user agent views the platform as the environment and the platform agent views the user as the environment. By this multi-agent view, \cite{shi2018virtual} proposed a multi-agent imitation method MAIL, extending the GAIL framework \cite{ho2016generative}, which learns the two policies simultaneously by beating the discriminator that finds the difference between the generated and the real interaction data.  %In this scenario, the policy of users could be learned from historical interactions by imitation learning approaches. In this way, the environment reconstruction is simplified to learn the policy of users, then we could conduct the traditional reinforcement learning in the built virtual environment. Generative Adversarial Network (GAN) is the milestone for the adversarial learning and recently Generative Adversarial Imitation Learning (GAIL) was proposed to take the advantage of the adversarial learning from the multi-agent interaction environment, deriving a model-free policy learning from data .

However, the MAIL method \cite{shi2018virtual} is under the assumption that the whole world consists of the two agents only. From the perspective of the real users, they can receive much more information from the real-world that is not recorded in the data. Therefore, it is still quite challenging to reconstruct the environment in real-world applications, since the real-world scenario is too complex to offer a fully observable environment, which means that it might exist the hidden confounders. As shown in Figure~\ref{fig:cg_rl}, in the classical setting, the next state depends on the previous state and the executed action. While in most of real-world scenarios, the next state could be extra influenced by some hidden confounders. If we follow the assumption of a fully observable world, the reconstruction may be misled by the appeared fake associations in the data, due to the unawareness of the possible hidden causes. Thus, it is essential to take hidden confounders into consideration.

Originally, confounder is a casual concept \cite{pearl2009causality}. It can affect both the treatment and the outcome in an experiment and cause a spurious association in observational data \cite{louizos2017causal}. Similarly, in reinforcement learning, hidden confounders can affect both actions and rewards as an agent interacts with the environment. When it comes to such real-world applications, it is necessary to involve the confounder into the learning task because of the confounding effect. Yet, little work has been done in this promising area \cite{forney2017counterfactual, bareinboim2015bandits}. To the best of our knowledge, this is the first study in reinforcement learning to reconstruct an environment together with hidden confounders. 

To involve hidden confounders into the environment reconstruction, we propose a \emph{deconfounded multi-agent environment reconstruction} method, named DEMER. Firstly, we formulate two representative agents, $\pi_a$ and $\pi_b$, interacting with each other. Then, in order to simulate the confounding effect of hidden confounders, we add a confounding agent $\pi_h$ into the formulation. According to the casual relationship, the confounding agent $\pi_h$ interacts with the other two agents. Based on the formulation, we learn each policy of three agents from the historical data by imitation learning. Since the hidden confounder is unobservable, to learn the policy of it, we propose two techniques: the confounder embedded policy and the compatible discriminator under the framework of GAIL \cite{ho2016generative}. The confounder embedded policy involves the confounding policy into the observable policy. The compatible discriminator is designed to discriminate the state-action pairs of the two observable policies so as to provide the respective imitation reward. As the training converges, the deconfounded environment is reconstructed. 

To verify the effectiveness of DEMER, we firstly use an artificial environment abstracted from the real application. Then, we apply DEMER to a large-scale recommender system for ride-hailing driver programs in Didi Chuxing. Through comparative evaluations, DEMER shows significant improvements in this real application.

The contribution of this work is summarized as follows:
\begin{itemize}
	\item We propose a novel environment reconstruction method to tackle the practical situation where hidden confounders exist in the environment. To the best of our knowledge, this is the first study to reconstruct environment with taking hidden confounders into consideration.
	\item By treating the hidden confounder as a hidden policy, we formulate the confounding effect into a multi-agent interactive environment. We propose an imitation learning framework by considering the interaction among two agents and the confounder. We define the confounder embedded policy and the compatible discriminator to learn policies effectively. 
	\item We deploy the proposed framework to the driver program recommendation system on a large-scale riding-hailing platform of Didi Chuxing, and achieve significant improvements in the test phase.
\end{itemize}

The rest of this paper is organized as follows: we introduce the background in Section~\ref{sec:bg} and the proposed method DEMER in Section~\ref{sec:DEMER}. We describe the application for the scenario of driver program recommendation in Section~\ref{sec:DEMERdar}. Experiment results are reported in Section~\ref{sec:exp}. Finally, we conclude the paper in Section~\ref{sec:ccs}.

\section{Reinforcement Learning and Environment Reconstruction}	\label{sec:bg}
\subsection{Reinforcement Learning}
The problem to be solved by Reinforcement Learning (RL) can usually be represented by a Markov Decision Processes (MDP) quintuple $ (S, A, T, R, \gamma) $, where $ S $ is the state space and $ A $ is the action space and $ T\colon S \times A \mapsto S $ is the state transition model and $ R\colon S \times A \mapsto \mathbb{R}  $ is the reward function and $ \gamma $ is the discount factor of cumulative reward. Reinforcement learning aims to optimize policy $ \pi\colon S \mapsto A $ to maximize the expected $ \gamma $-discounted cumulative reward $ \mathbb{E}_\pi[\Sigma_{t=0}^T\gamma^tr_t] $
by enabling agents to learn from interactions with the environment. The agent observes state $ s $ from the environment, selects action $ a $ given by $ \pi $ to execute in the environment and then observes the next state, obtains the reward $ r $ at the same time until the terminal state is reached. Consequently, the goal of RL is to find the optimal policy
\begin{equation}
	\centering
	\pi^\star = \mathop{\arg\max}_\pi \mathbb{E}_\pi[\Sigma_{t=0}^T \gamma^tr_t] \,,
\end{equation}
of which the expected cumulative reward is the largest. 

\textbf{Imitation Learning}. Learning a policy directly from expert demonstrations has been proven very useful in practice, and has made a significant improvement of performance in a wide range of applications \cite{ross2011reduction}. There are two traditional imitation learning approaches: behavioral cloning, which trains a policy by supervised learning over state-action pairs of expert trajectories \cite{pomerleau1991efficient}, and inverse reinforcement learning \cite{russell1998learning}, which learns a cost function that prioritizes the expert trajectories over others. Generally, common imitation learning approaches can be unified as the follow formulation: training a policy $ \pi $ to minimize the loss function $ l(s, \pi(s)) $, under the discounted state distribution of the expert policy: $ P_{\pi_e}(s) = (1-\gamma)\Sigma_{t=0}^T \gamma^t p(s_t) $. The object of imitation learning is represented as 
\begin{equation}
	\label{eq:irl}
	\centering
	\pi = \mathop{\arg\min}_{\pi} \mathbb{E}_{s\sim P_{\pi_e}} [l(s, \pi(s))] \,,
\end{equation}

\textbf{Confounding Reinforcement Learning}. Originally, confounding is a concept in casual inference \cite{pearl2009causality}. Confounder is a variable that influences both the treatment and the outcome, naturally corresponding to the action and the reward in reinforcement learning. From the perspective of traditional reinforcement learning, the state is a confounder between the action and the reward. Although there are inherent similarities between causal inference and reinforcement learning, little work has been done in reinforcement learning that confounders exist in the environment \cite{forney2017counterfactual, bareinboim2015bandits}. 
Only recently, Lu et al.~\cite{lu2018deconfounding} proposed the deconfounding reinforcement learning to adapt to the confounding setting, while the model of confounder is stationary at each time step which actually can be dynamic.

\subsection{Environment Reconstruction}
Reinforcement learning relies on an environment. However, when it comes to real-world applications, it is not practical to interact with the real-world environment directly to optimize the policy because of the low sampling efficiency and the high-risk uncertainty, such as online recommendation in E-commerce and medical diagnosis. A viable option is to reconstruct a virtual environment \cite{shi2018virtual}. As a result, the learning process could be more efficient by interacting with the virtual environment and the interaction cost could be avoided as well.

\textbf{Generative Adversarial Nets}.
Generative adversarial networks (GANs) \cite{goodfellow2014generative} and its variants are rapidly emerging unsupervised machine learning techniques. GANs involve training a generator $G$  and discriminator $D$ in a two-player zero-sum game:
\begin{equation}
	\centering
	\mathop{\arg\min}_G \mathop{\arg\max}_{D\in(0,1)} \mathbb{E}_{x \sim p_E}[ \log D(x)] + \mathbb{E}_{z\sim p_z}[\log(1-D(G(z)))] \,,
\end{equation}
where $ p_z $ is some noise distribution. In this game, the generator learns to produce samples (denoted as $ x $ ) from a desired data distribution (denoted as $ p_E $). The discriminator is trained to classify the real samples and the generated samples by supervised learning, while the generator $G$ aims to minimize the classification accuracy of $D$ by generating samples like real ones. In practice, the discriminator and the generator are both implemented by neural networks, and updated alternately in a competitive way. The training process of GANs can be seen as searching for a Nash equilibrium in a high-dimensional parameter space. Recent studies have shown that GANs are capable of generating faithful real-world images \cite{menick2018generating}, demonstrating their applicability in modeling complex distributions.

\textbf{Generative Adversarial Imitation Learning}. GAIL \cite{ho2016generative} has become a popular imitation learning method recently. It was proposed to avoid the shortcoming of traditional imitation learning, such as the instability of behavioral cloning and the complexity of inverse reinforcement learning. It adopts the GAN framework to learn a policy (i.e., the generator $G$) with the guidance of a reward function (i.e., the discriminator $D$) given expert demonstrations as real samples. GAIL formulates a similar objective function like GANs, except that here $ p_E $ stands for the expert's joint distribution over state-action pairs:
\begin{equation}
	\centering
	\mathop{\arg\min}_{\pi} \mathop{\arg\max}_{D\in(0,1)} \mathbb{E}_{\pi}[ \log D(s,~a)] + \mathbb{E}_{\pi_E}[\log(1-D(s,~a))] - \lambda H(\pi) \,,
\end{equation}
where $ H(\pi) \triangleq \mathbb{E}_{\pi}[-\log \pi(a|s)] $ is the entropy of $ \pi $.

GAIL allows the agent to execute the policy in the environment and update it with policy gradient methods \cite{schulman2015trust}. The policy is optimized to maximize the similarity between the policy-generated trajectories and the expert ones measured by $D$. Similar to the equation (\ref{eq:irl}), the policy $ \pi $ is updated to minimize the loss function
\begin{equation}
\label{eq:gail}
l(s, \pi(s)) = \mathbb{E}_{\pi}[ \log D(s, a) ] - \lambda H(\pi) \cong \mathbb{E}_{\tau_i}[\log\pi(a|s)Q(s,a)] -\lambda H(\pi)\,.
\end{equation}
where $ Q(s, a) = \mathbb{E}_{\tau_i} [\log (D(s, a))| s_0=s, a_0=a]$ is the state-action value function. 
The discriminator is trained to predict the conditional distribution: $ D(s,a)=p(y|s,a) $ where $ y \in \{\pi_E, \pi\} $. In other words, $ D(s,a) $ is the likelihood ratio that the pair $ (s,a) $ comes from $ \pi $ rather than from $ \pi_E $. GAIL is proven to achieve similar theoretical and empirical results as IRL \cite{finn2016connection} while it is more efficient. Recently, the multi-agent extension of GAIL \cite{shi2018virtual} has been proven effective to reconstruct an environment.

Shi et al.~\cite{shi2018virtual} proposed to virtualize an online retail environment by extending the GAIL framework to a multi-agent approach, MAIL, that learns the interacting factors simultaneously. They showed that the multi-agent method leads to a better generalizable environment. 

\section{Deconfounded Multi-agent Environment Reconstruction}	\label{sec:DEMER}
To reconstruct environments where hidden confounders exist, we propose a novel \emph{deconfounded multi-agent environment reconstruction} (DEMER) method.

In this study, by treating the hidden confounder as a hidden policy, we formulate the deconfounding environment reconstruction as follows: there are two agents $ A $ (known as the policy agent) and $ B $ (known as the environment), interacting with each other and both of them are confounded by a hidden confounder $ H $. Specifically, the dynamic effect of the hidden confounder $ H $ is modeled as a hidden policy $ \pi_h $. The observation and action of each agent are defined as follows: Given $ o_A $ as the observation of agent $ A $, it takes an action $ a_A = \pi_a(o_A)$. The observation $ o_H $ of the hidden policy is formatted as the concatenation $ o_H = <o_A, a_A> $, and action $ a_H = \pi_h(o_H) $ has the same format as $ a_A $. The observation $ o_B $ of agent $ B $ is formatted as the concatenation $ o_B = <o_A, a_A, a_H> $, and its action is $ a_B = \pi_b(o_B) $ which can be used to move forward to the next state. The objective is to use only observed interactions, that is, trajectories $ \{(o_A, a_A, a_B)\} $, to imitate the policies of $ A, B $ and recover the potential effect of $ H $ by inferring the hidden policy $ \pi_h $. 
The objective function of multi-agent imitation learning is then defined analogy to equation (\ref{eq:irl}):
\begin{equation}
\centering
(\pi_a, \pi_b, \pi_h) = \mathop{\arg\min}_{(\pi_a, \pi_b, \pi_h)} \mathbb{E}_{s\sim P_{\tau_{real}}} [L(s, a_A, a_B)] \,,
\end{equation}
where $ a_A,~a_B $ depend on three policies. By adopting the GAIL framework, according to equation (\ref{eq:gail}), we can get
\begin{equation}
\label{eq:lsa}
    L(s,a_A, a_B) = \mathbb{E}_{\pi_a, \pi_h, \pi_b}[\log D(s, a_A, a_B)]-\lambda \Sigma_{\pi \in \{\pi_a, \pi_h, \pi_b\}}H(\pi)
\end{equation}
and observe that $\pi_a$ is independent with $\pi_h$ and $\pi_b$ given $s$ and $a_A$, then
\begin{equation}
\label{eq:dsa}
\begin{aligned}
D(s, a_A, a_B)  &= p(\pi_a, \pi_h, \pi_b|s,a_A,a_B)  \\
&= p(\pi_a |s,a_A,a_B)~p( \pi_h, \pi_b|s,a_A,a_B) \\
&= p(\pi_a |s,a_A)~p( \pi_h, \pi_b|s,a_A,a_B)  \\
&= D_a(s, a_A)~D_{hb}(s,a_A, a_B) \,.
\end{aligned}
\end{equation}
Combining equations (\ref{eq:lsa}) and (\ref{eq:dsa}), we can get the formulation as
\begin{equation}
\label{eq:demer}
\begin{aligned}
L(s,a_A, a_B) = &~\mathbb{E}_{\pi_a, \pi_h, \pi_b}[\log D_a(s, a_A) D_{hb}(s,a_A, a_B)] - \\ &~\lambda \Sigma_{\pi \in \{\pi_a, \pi_h, \pi_b\}}H(\pi) \\
= &~\mathbb{E}_{\pi_a}[\log D_a(s, a_A)] - \lambda H(\pi_a) + \\&~\mathbb{E}_{\pi_h, \pi_b}[\log D_{hb}(s,a_A, a_B)] -\lambda \Sigma_{\pi \in \{\pi_h, \pi_b\}}H(\pi)\\
= &~l(s, \pi_a(s)) + l((s, a_A), \pi_b \circ \pi_h((s,a_A)))
\end{aligned}
\end{equation}
which indicates that the optimization can be decomposed as optimizing policy $ \pi_a $ and joint policy $\pi_{hb} = \pi_b \circ \pi_h $ individually by minimizing
\begin{equation}
\label{eq:demer_a}
\begin{aligned}
	l(s, \pi_a(s)) &= \mathbb{E}_{\pi_a}[\log D_a(s, a_A)] - \lambda H(\pi_a) \\
				   &\cong \mathbb{E}_{\tau_i}[\log\pi_a(a_A|s)Q(s,a_A)] -\lambda H(\pi_a) \,,
\end{aligned}
\end{equation}
where  $ Q(s, a_A) = \mathbb{E}_{\tau_i} [\log (D(s, a_A))| s_0=s, a_0=a_A]$ is the state-action value function of $ \pi_a $, and 

\begin{equation}
\label{eq:demer_hb}
\begin{aligned}
l((s, a_A), \pi_{hb}((s,a_A))) = &~\mathbb{E}_{\pi_h, \pi_b}[\log D_{hb}((s,a_A), a_B)] -\\ &~\lambda \Sigma_{\pi\in\{\pi_h,\pi_b\}} H(\pi) \\
\cong &~\mathbb{E}_{\tau_i}[\log \pi_{hb}(a_B|s, a_A)Q(s,a_A, a_B)] - \\
&~\lambda \Sigma_{\pi\in\{\pi_h,\pi_b\}} H(\pi) \,,
\end{aligned}
\end{equation}
where  $ Q(s, a_A, a_B) = \mathbb{E}_{\tau_i} [\log (D((s, a_A), a_B))| s_0=s, a_{A0}=a_A, a_{B0}]$ is the state-action value function of $ \pi_{hb} $.

Based on this result, we propose the confounder embedded policy and the compatible discriminator to achieve the goal of imitating polices of each agent, thus obtaining the DEMER approach.

\subsection{Confounder Embedded Policy}
In this study, the interaction between the agent $A$ (known as the policy agent) and the agent $B$ (known as the environment) could be observed, while the policy and data of the agent $H$ (known as hidden confounders) are unobservable. Thus, we combine the confounder policy $ \pi_h $ with policy $ \pi_b $ as a joint policy named $ \pi_{hb} = \pi_b \circ \pi_h $. Together with the policy $ \pi_a $, the generator is formalized as an interactive environment of two policies as shown in the top of Figure~\ref{fig:pi_joint}. The joint policy can actually be expressed as $ \pi_{hb} (o_A, a_A) = \pi_b(o_A, a_A, \pi_h(o_A, a_A)) $ in which the input $ o_A, a_A $ and the output $ a_B $ are both observable from the historical data. Therefore, we can use imitation learning methods to train these two policies by imitating the observed interactions.

The policies in generator are updated respectively for each training step: firstly the joint policy $ \pi_{hb} $ is updated with the reward $ r^{HB} $ given by the discriminator, secondly the policy $ \pi_a $ is updated with the reward $ r^A $. Though there is no explicit updating step for the hidden confounder policy $ \pi_h $, it has been optimized iteratively by these two steps. Intuitively, the generated hidden policy $ \pi_h $ is just like a by-product along with the process of optimizing policies $ \pi_a $ and $ \pi_{hb} $ towards the truth and in consequence it can recover the real confounding effect to some extent. To make the training process more stable, we employ TRPO to update policies mentioned above.

\subsection{Compatible Discriminator}
\begin{figure}[t]
	\centering
	\includegraphics[width=\linewidth]{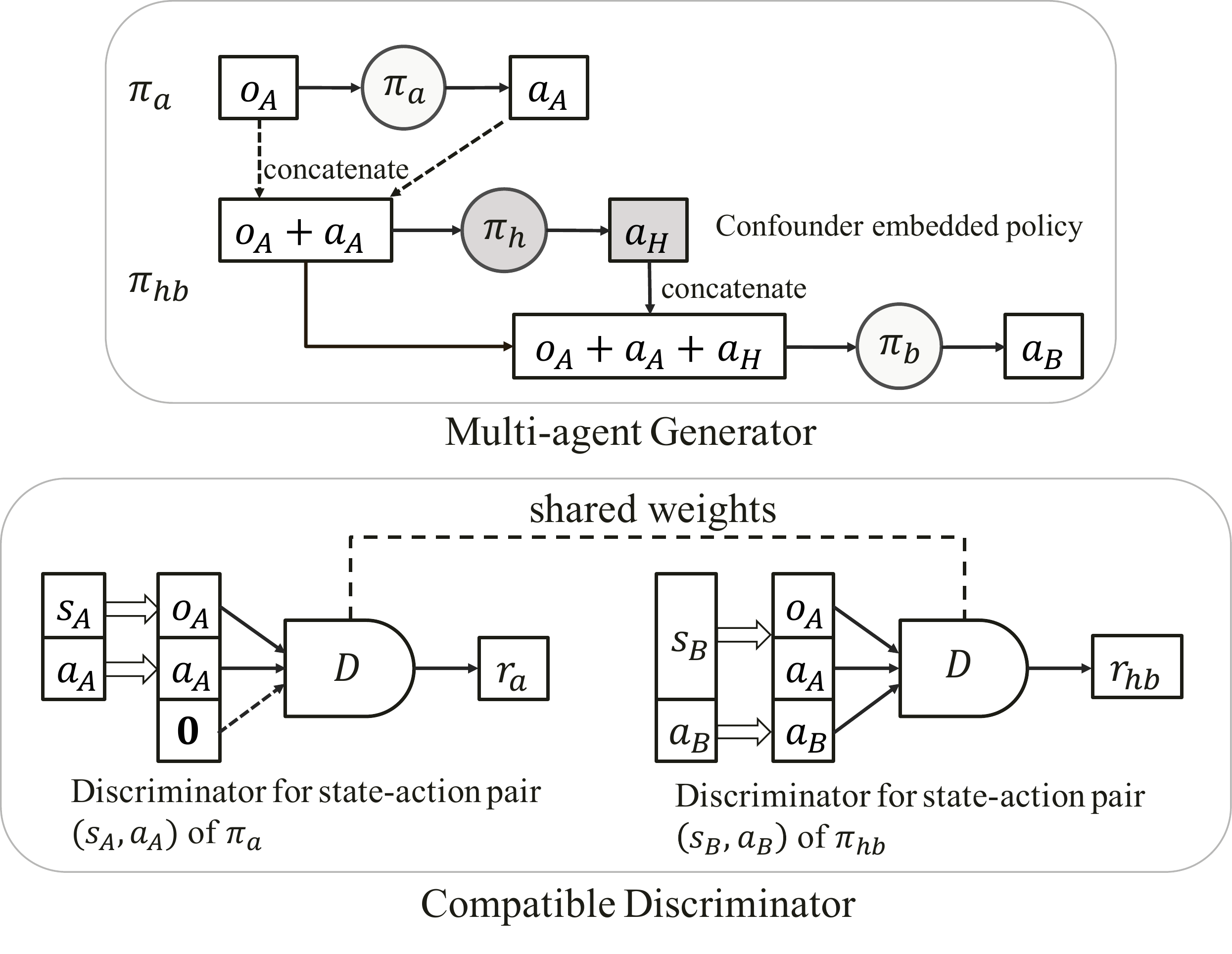}
	\caption{The generator and discriminator in DEMER.}
	\label{fig:pi_joint}
\end{figure}

In most of generative adversarial learning frameworks, there is only one task to model and learn in the generator. In this study, it is essential to simulate and learn different reward functions for the two policies $ \pi_a, \pi_{hb} $ in the generator respectively. Thus, we design the discriminator compatible with two classification tasks. As Figure~\ref{fig:pi_joint} illustrates, one task is designed to classify the real and generated state-action pairs of $ \pi_{hb} $ while the other one is to classify the state-action pair of $ \pi_a $. Correspondingly, the discriminator has two kinds of input: the state-action pair $ (o_A,~a_A,~a_B) $ of policy $ \pi_{hb} $ and the zero-padded state-action pair $ (o_A,~a_A,~\mathbf{0}) $ of policy $ \pi_a $. This setting indicates that the discriminator splits not only the policy $ \pi_{hb} $'s state-action space, but also the policy $ \pi_a $'s. The loss function of each task is defined as
\begin{equation}
\label{eq:dloss_hb}
	 E_{\tau_{sim}}[\log(D_\sigma(o_A,a_A,a_B))]+E_{\tau_{real}}[\log(1-D_\sigma(o_A,a_A,a_B))] 
\end{equation}
for $ \pi_{hb} $ , and
\begin{equation}
\label{eq:dloss_a}
 	E_{\tau_{sim}}[\log(D_\sigma(o_A,a_A,\mathbf{0}))]+E_{\tau_{real}}[\log(1-D_\sigma(o_A,a_A,\mathbf{0}))]
\end{equation}
for policy $ \pi_a $. 

The output of the discriminator is the probability that the pair data comes from the real data.  The discriminator is trained with supervised learning by labeling the real state-action pair as $1$ and the generated fake state-action pair as $0$. Then it is used as a reward giver for the policies while simulating interactions. The reward function for policy $ \pi_{hb} $ can be formulated as:
\begin{equation}	\label{eq:r_hb}
r^{HB} = -\log(1-D(o_A,~a_A,~a_B)) \,,
\end{equation}
and the reward function for policy $ \pi_a $ is 
\begin{equation}	\label{eq:r_a}
r^A = -\log(1-D(o_A,~a_A,~\mathbf{0})) \,.
\end{equation}

\subsection{Simulation}
We simulate interactions in the generator module. The policy-generated trajectory is generated as follows: Firstly, we randomly sample one trajectory from the observed data and set its first state as the initial observation $ o_0^A $. Then we can use the two policies $ \pi_a, \pi_{hb} $ to generate a whole trajectory triggered from $ o_0^A $. Given the observation $ o_t^A $ as the input of $ \pi_a $, the action $ a_t^A $ can be obtained. In consequence, the action $ a_t^B $ can be obtained from the joint policy $ \pi_{hb} $ with the concatenation $ <o_t^A, a_t^A> $ as input. Then we can get the simulation reward $ r_t^{HB} $ by equation~(\ref{eq:r_hb}) and $ r_t^A$ by equation~(\ref{eq:r_a}) which would be used for updating policies in the adversarial training step. Next, we can get the next observation $ o_{t+1}^A $ given $ o_t^A $ and $ a_t^B $ by the predefined transition dynamics. This step is repeated until the terminal state and a fake trajectory is generated.

\subsection{DEMER Algorithm}
Based on the confounder embedded policy and the compatible discriminator, we propose the DEMER method to achieve the goal of reconstructing environment with hidden confounders from the observed data.

\begin{algorithm}[t]
	\floatname{algorithm}{Algorithm}
	\renewcommand{\algorithmicrequire}{\textbf{Input:}}
	\caption{DEMER}
	\label{alg:DEMER}
	\begin{algorithmic}[1]
		\Require $ D_{real}=\{\tau_1, \tau_2, \ldots, \tau_n\} \colon$ The observed real trajectories over $ T $ steps;
		\Statex ~~~~~~$ N \colon$ Number of trajectories generated in each iteration;
		\Statex ~~~~~~$ K \colon$ Steps of generator per discriminator step;
		\State Initialize parameters $ \theta^{hb} $ and $ \theta^a $ of policy $ \pi_{hb} $ and $ \pi_{a} $,  parameters $ \sigma $ of discriminator $ D $;
		\For{$ i = 1, 2, \ldots $} 
		\For{$ k = 1, 2, \ldots, K $}
		\State $ \tau_{sim} = \varnothing $ ;
		\For{$ j = 1, 2, \ldots, N $}
		\State $\tau_j = \varnothing $ ;
		\State Randomly sample one trajectory $ \tau_r $ from $ D_{real} $ and set its first state as the initial observation $ o_0^A $ ;
		\For{$ t = 0, 2, \ldots, T-1 $}
		\State Simulate the action $ a_t^A = \pi_a(o_t^A) $ ;
		\State Simulate the action $ a_t^B = \pi_{hb}(o_t^A, a_t^A) $ ;
		\State Get reward $ r_t^A $ according to Equation~(\ref{eq:r_a}) ;
		\State Get reward $ r_t^{HB} $ according to Equation~(\ref{eq:r_hb}) ;
		\State Get next observation $ o_{t+1}^A $ given $ o_t^A,~a_t^B $ by predefined transition;
		\State Add $\{o_t^A,~a_t^A,~a_t^B,~r_t^A,~r_t^{HB}\} $ to $\tau_j$;
		\EndFor
		\State Add $ \tau_j $ to $ \tau_{sim} $ ;
		\EndFor
		\State TRPO update $ \theta^{a} $ and $ \theta^{hb}  $ with simulation trajectories $ \tau_{sim} $ according to the equation (\ref{eq:demer_a}) and (\ref{eq:demer_hb}) respectively;
		\EndFor
		\State Update the parameters $ \sigma $ of the discriminator $ D $ by minimizing the losses in equation (\ref{eq:dloss_hb}) and (\ref{eq:dloss_a}) in turn;
		\EndFor
		\State \Return the trained policies $ \pi_a, ~\pi_b, ~\pi_h $ .
	\end{algorithmic}
\end{algorithm}

Algorithm~\ref{alg:DEMER} shows the details of DEMER. The whole algorithm adopts the generative adversarial training framework. In each iteration, firstly the generator simulates interactions using policies $ \pi_a,~\pi_{hb} $ to collect the trajectory set $ \tau_{sim} $ corresponding to the line 5 to line 17. Then the policy $ \pi_{a} $ and $ \pi_{hb} $ are updated in turn using TRPO with generated trajectories $ \tau_{sim} $ in line 18. After $ K $ generator steps, the compatible discriminator is trained by two steps as shown in line 20. Specifically, the predefined transition dynamics in line 13 depends on specific tasks. The DEMER method can effectively imitate the policies of observed interactions and recover the hidden confounder beyond observations.

\section{Application in Driver Program Recommendation}	\label{sec:DEMERdar}

\begin{figure*}[t]
	\centering
	\begin{minipage}{0.26\textwidth}
	\caption{DEMER framework applied in the driver program recommendation. While real-world data only collects the interactions between the drivers and the Didi Chuxing platform, the virtual environment contains three policies simulating the drivers, the platform, and the confounding variable.}
	\label{fig:scenario}
	\end{minipage}\hfill
	\begin{minipage}{0.65\textwidth}
	\includegraphics[width=\textwidth]{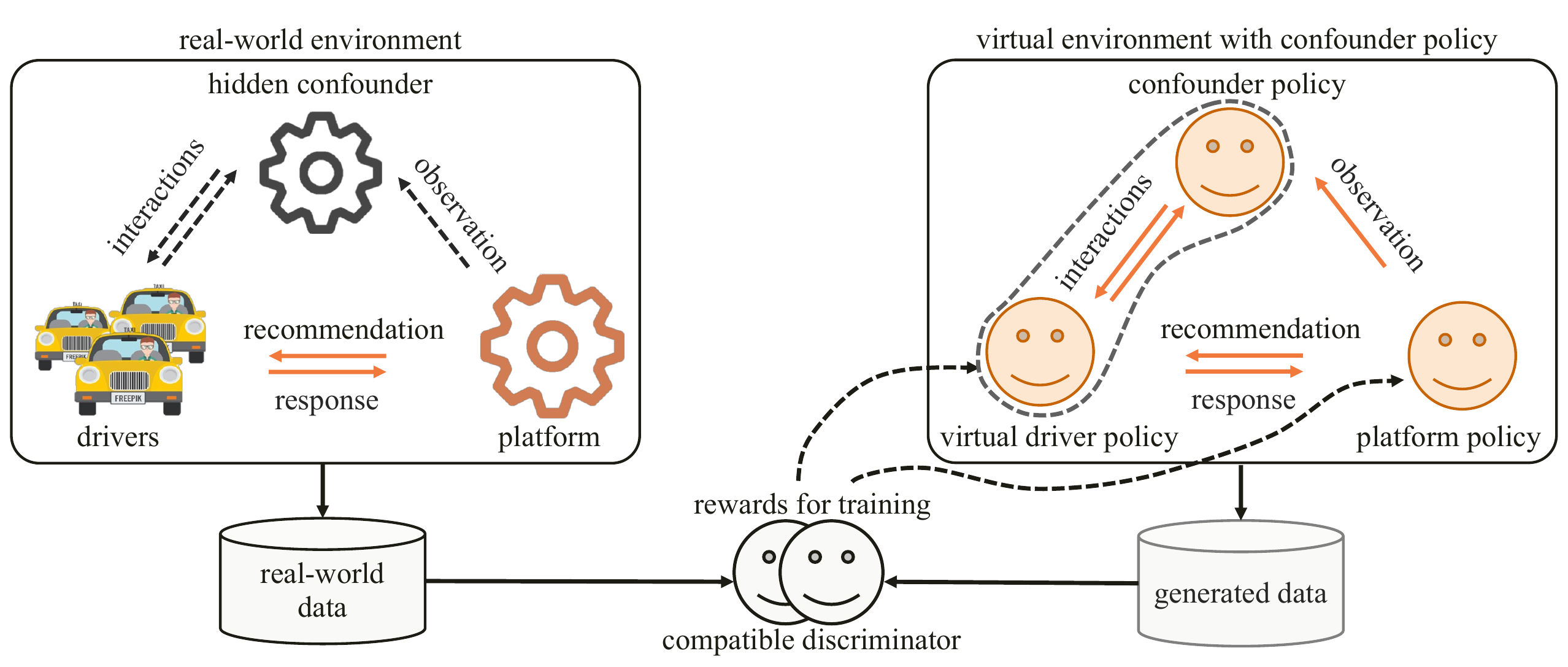}
	\end{minipage}
\end{figure*}

\subsection{Driver Program Recommendation}	\label{sec:dar}
We have witnessed a rapid development of on-demand ride-hailing services in recent years. In this economic pattern, the platform often recommends programs to drivers, aimed to help them finish more orders. Specifically, the platform would select the appropriate program to recommend the drivers to participate every day, and then adjust the program content according to the drivers' feedback behavior. This is a typical reinforcement learning task. However, since the behavior of drivers is not only affected by the recommended programs, but also affected by some other unobservable factors, such as the response to special events and so on, that is, hidden confounders. In order to achieve the goal of policy optimization, it is essential to take into account the potential influence of hidden factors when recommending programs.

However, traditional reinforcement learning approaches are applied in these problems without exploring the impact of hidden confounders, which would consequently degrade the learning performance. Thus, a more adaptive approach such as DEMER proposed in this paper is desirable to tackle these problems. 

\subsection{DEMER based Driver Program Recommendation}
As for the driver program recommendation, we apply DEMER to build a virtual environment with hidden confounders by using historical data. As shown in Figure~\ref{fig:scenario}, there are three agents in the environment, representing driver policy $ \pi_d $, platform policy $ \pi_p $ and confounder policy $ \pi_h $. We can see that the driver policy and the platform policy have the nature of ``mutual environment" from the perspective of MDP. From the platform's point of view, its observation is the driver's response, and its action is the recommendation program to the driver. Correspondingly, from the driver's point of view, its observation is the platform's recommendation program, and its action is the driver's response to the platform. The hidden confounder is modeled as a hidden policy same as DEMER, so as to make a dynamic effect at each time step. 

\textbf{Data preparation}. 
Based on the above scenario, we integrate the historical data and then construct historical trajectories $ D_{hist} =  \{ \tau_1, \ldots, \tau_i, \ldots, \tau_n \} $ representing trajectories of $ n $ drivers. Each trajectory $ \tau_i = \{o_0^P, a_0^P, a_0^D, o_1^P, \ldots, o_t^P, a_t^P, a_t^D, o_{t+1}^P, \ldots, o_T^P \} $ represents the T steps of interactions of driver $ d_i $. 

\textbf{Definition of policies}. 
According to the interaction among agents in this scenario, the observation and action of each agent policy are defined as follows:
\begin{itemize}
	\item {platform policy $ \pi_p $}: The observation $ o_t^P $ consists of the driver's static characteristics (using real data) and the simulated response behavior $ a_{t-1}^D $; the action $ a_t^P $ is the program information recommended for the driver.
	\item {hidden policy $ \pi_h $}: The observation $ o_t^H $ consists of $ o_t^P $ and $ a_t^P $; the action $ a_t^H $ is the same format as  $ a_t^P $.
	\item {driver policy $ \pi_d $}: The observation $ o_t^D $ consists of $ o_t^P $,  $ a_t^P $ and $ a_t^H $; the action $ a_t^D $ is the simulated driver's behavior at the current step.
\end{itemize}
Similar to the DEMER setting, we further combine the policies $ \pi_h,~\pi_d $ into a joint policy named $ \pi_{joint} $. We then apply DEMER to train $ \pi_{joint} $ and $ \pi_p $. Afterwards, the deconfounding environment of driver program recommendation is reconstructed.

\textbf{ RL in virtual environment}. 
Once the deconfounding virtual environment is built, we perform reinforcement learning efficiently to optimize the policy $ \pi_p $ by interacting with the environment. Due to the simulated confounders in the environment, the reinforcement learning approach could learn a deconfounding policy with improved performance in the real world.

\section{EXPERIMENTS}	\label{sec:exp}
In this section, we conduct two groups of experiments to validate the effect of DEMER method. One is a toy experiment in which we design an environment with predefined rules, the other is a real-world application of driver program recommendation on a large-scale ride-hailing platform Didi Chuxing.

\subsection{Artificial Environment}
%In order to evaluate the effectiveness of DEMER to recover the unobservable confounding function and the observable policy functions, 
We firstly hand-craft an artificial environment, consisting of the artificial platform policy $\pi_p$, the artificial driver policy $\pi_d$, and the artificial confounder $\pi_h$, with deterministic rules to mimic the real environment. Then we use  DEMER to learn the policies and compare with the real rules. Besides, we conduct the MAIL method, without modeling hidden confounders, as a comparison.

\begin{figure}[t]
	\centering
	\includegraphics[width=0.8\linewidth]{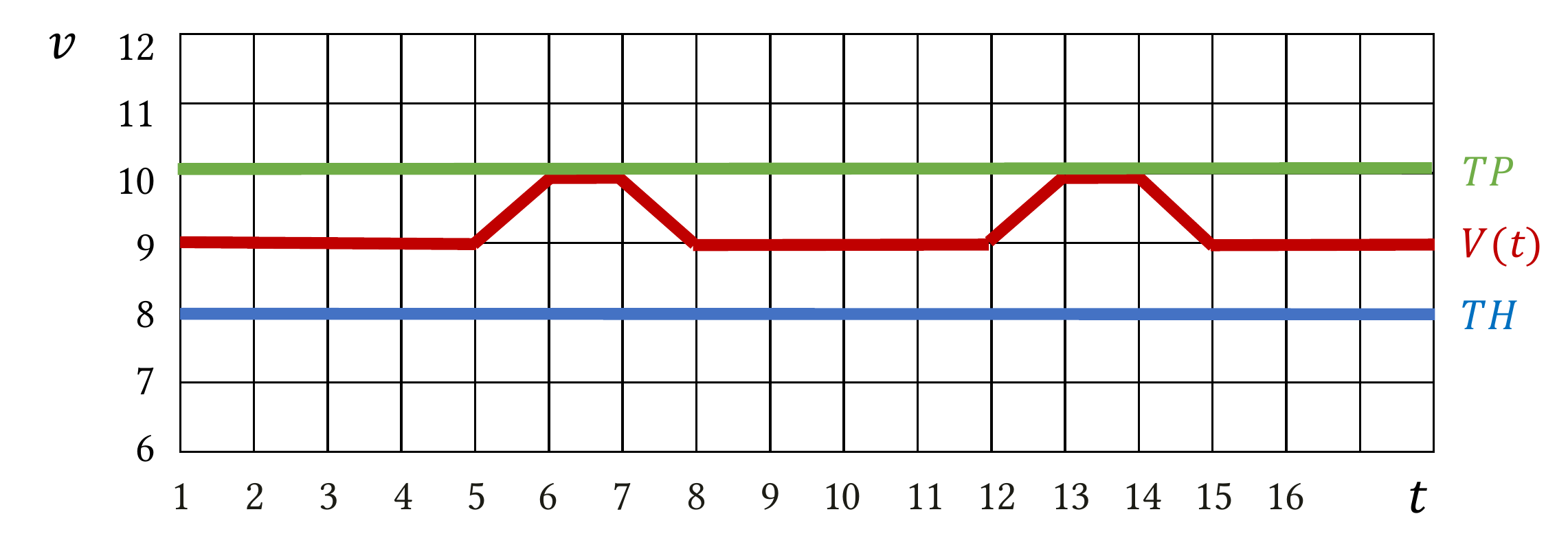}
	\caption{Schematic drawing of interaction in the toy environment: $t$ represents the time step and $v$ is a variant affected by all three policies. $TP$ and $TH$ are the thresholds for policies taking effect, and $V(t)$ describes the intrinsic evolution trend of the artificial driver policy $\pi_d$.}
	\label{fig:tedraft}
\end{figure}

\textbf{Description of the artificial environment}.
Similar to the interaction in the scenario of driver program recommendation, we define a triple-agents environment to simulate a Markov decision process. The semantic drawing of this toy scenario is shown in Figure~\ref{fig:tedraft}. In a Markov decision process, the key variant $v$ (denotes the driver's response) is affected by three policies at each time step. The policy $\pi_d$ has an intrinsic evolution trend on the variant $v$ in the period of 7 time steps, as defined in equation~(\ref{eq:deltaV}). The policy $\pi_p$ has a positive effect on the variant $v$ if the value of $v$ is under the green line else no effect. Oppositely, the policy $\pi_h$ has a negative effect on the variant $v$ if the value of $v$ is above the blue line else no effect. The green and blue lines can be seen as the thresholds of $\pi_p$ and $\pi_h$ to make effect on the evolution trend of $v$. Here we set the policy $\pi_h$ as a role of hidden confounders in this environment, of which the effect on the interaction would not be observed.

\textbf{MDP definition}.
The \textit{observation} $o$ is a tuple $(tw, r, v)$, in which $tw \in \{1, 2, \ldots, 7\}$ is the time step in one period, $ r $ is a static factor used to make a difference on the effect of each agent and $v$ is the key variant in the interaction process. The initial value $v_0$ is sampled from a uniform distribution $U(9+wave, 9-wave), wave = 1.2$. We add the static factor $r = 1-0.5\times\frac{v_0-9}{wave}$ into the state to make the episodes generated by this setting more diverse. 

The \textit{action} is defined as the output of the deterministic policy. The thresholds of green line $TP$ and blue line $TH$ are 10 and 8 correspondingly. Then we define the deterministic policy rule of each agent as follows:
\begin{align}
	&a_p = \pi_p(tw, r, v) = \max(0, \min(1, r\times(TP - v) \times \frac{tw}{7})) \label{eq:pi_p} \,, \\
	&a_h = \pi_h(tw, r, v, a_p) = \max(-1, \min(0, r\times(TH - v - \frac{a_p}{2}) \times \frac{tw}{7})) \label{eq:pi_h} \,,\\
	&a_d = \pi_d(tw, r, v, a_p, a_h) = \Delta V(tw) + a_p +a_h \label{eq:pi_d}  \,.
\end{align}
where
\begin{equation}	\label{eq:deltaV}
\Delta V(tw) = \begin{cases}
1 & \text{if $tw = 5$;}\\
-1 & \text{else if $tw = 7$;}\\
0 & \text{otherwise.}
\end{cases}
\end{equation}

The \textit{transition dynamics} is simply defined as: $ v_{t+1} = v_t + a_d^t $ and $ r $ is a constant once initialized. $tw$ is a timestamp indicator cycling in the sequence $\left[  1, 2, \ldots, 7 \right]$. In this experiment, we set the length of trajectory $T$ to 8.

By running in the toy environment, we collect many episodes as training data. Each episode is formatted as $\{o_p^0, a_p^0, a_d^0, o_p^1, \ldots, o_p^T\}$. Note that there is no action of policy $\pi_h$ in the episode.

\textbf{Implementation details}.
We conduct two training settings on this artificial environment: DEMER and MAIL. The major difference is that there is no confounding policy in the MAIL setting. We aim to compare the similarity between the generated policies and the defined rules. In detail, each policy is embodied by a neural network with 2 hidden layers and combined sequentially into a joint policy network illustrated in Figure~\ref{fig:pi_joint}. There are $64$ neurons in each hidden layer activated by $\tanh$ functions. To control the same complexity of the policy model, the joint policy network in MAIL has the same number of hidden layers as DEMER. The discriminator network adopts the same structure as each policy network. Different from GANs training, we perform $K = 3$ generator steps per discriminator step, and sample $N = 200$ trajectories per generator step. The detail of the training process is described in Section~\ref{sec:DEMER}. 

\begin{figure*}[t]
	\centering
	\includegraphics[width=0.8\linewidth]{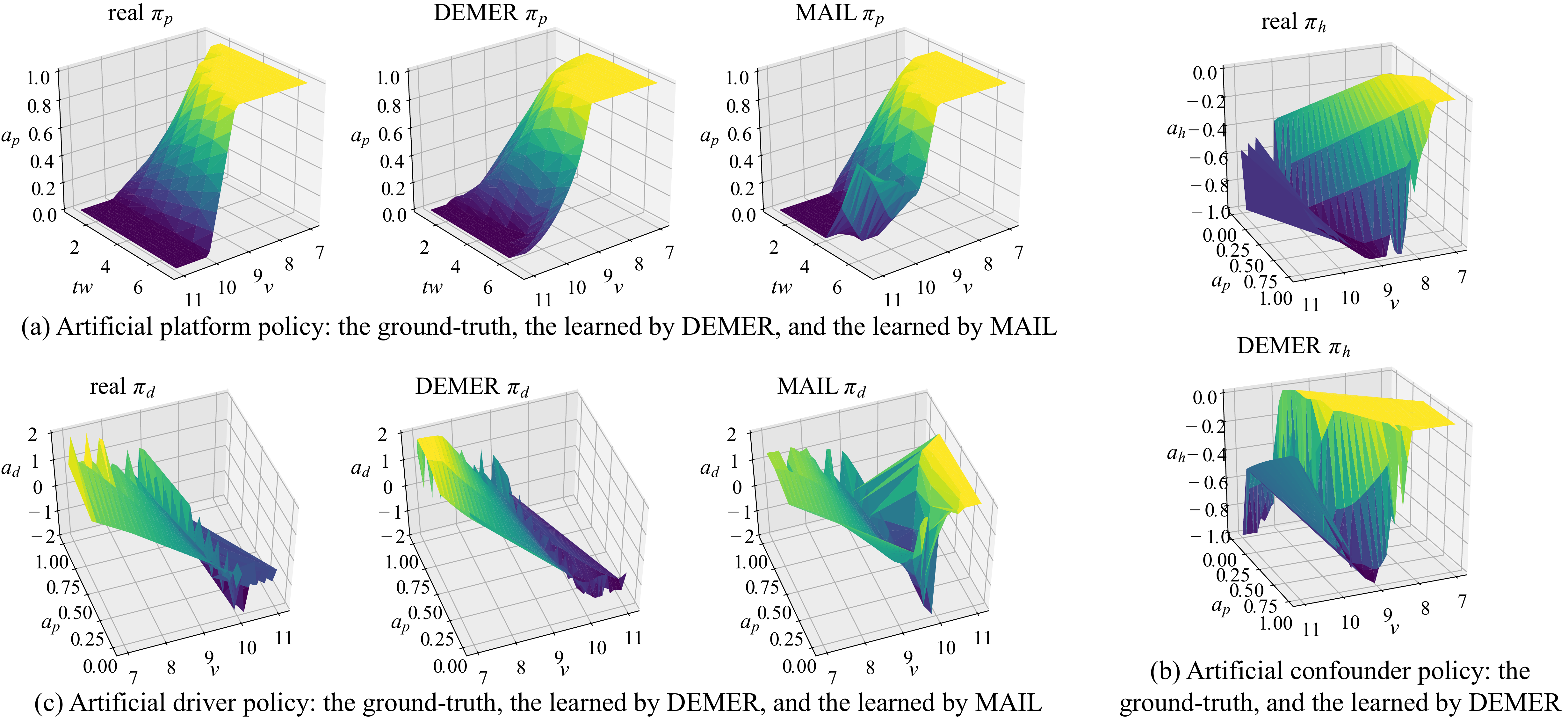}
	\caption{Visualization and comparison of policy functions, with $r=1.3$. More visualizations with various of $r$ values are presented in the supplement material.}
	\label{fig:toy_all}
\end{figure*}

\textbf{Results}.
The generated policy functions trained by DEMER and MAIL are shown in Figure~\ref{fig:toy_all}. 

First of all, from the perspective of the two observable policies, the policy function maps of $ \pi_p $ and $ \pi_d $ produced by DEMER are both more similar to the real function space than those by MAIL, as shown in Figures~\ref{fig:toy_all} [a] and [c]. MAIL produces sharp distortion shape locally when $r$ is large. We believe that this is because the hidden confounder has a greater impact on the interaction as $r$ increases, and a large confounding bias has reached a point where it cannot be neglected. 

Then we further compare the similarity between the confounder policy generated by DEMER and the true policy $\pi_h$. In Figure~\ref{fig:toy_all} [b], it can be seen the generated confounder policy can describe threshold effects well and match the real function map roughly, although it is difficult under the setting of fully unobservable confounders. Our results show the great potential of using observational data to infer the hidden confounder model.

\subsection{Real-world Experiment}
In this part, we apply DEMER to a real-world scenario of driver program recommendation as introduced in Section~\ref{sec:dar}. Firstly, we use historical data to reconstruct four virtual environments by four comparative methods. Next, we evaluate these environments from various statistical measures. Finally, we train four recommendation policies in these environments by the same training method and evaluate these policies in offline and online environments.

Specifically, we include four methods in our comparison: 
\begin{itemize}
	\item{\textbf{SUP}}: Supervised learning of the driver policy with historical state-action pairs, i.e., behavioral cloning; 
	\item{\textbf{GAIL}}: GAIL to learn the driver policy, given the historical record of program recommendation as a static environment;
	\item{\textbf{MAIL}}: Multi-agent adversarial imitation learning, without modeling the hidden confounder.
	\item{\textbf{DEMER}}: The proposed method in this study;
\end{itemize}
We evaluate the models by different statistical metrics.

\textbf{Log-likelihood of real data on models}.
We evaluate the policy distribution of four different models by the mean log-likelihood of real state-action pairs on both training set and testing set. As shown in Table~\ref{tab:logll}, the model trained by DEMER achieves the highest mean log-likelihood on both data sets. Since the evaluation is on the view of each state-action pair, the behavioral cloning method SUP achieves a better performance than MAIL. While our method DEMER makes a significant improvement on MAIL, which indicates the positive influence of our confounder setting.

\textbf{Correlation of key factors trend}.
Another important measurement of generalization performance is the trend of drivers' response. We use two indicators' trend lines to compare different simulators: Number of finished orders (FOs) and Total Driver Incomes (TDIs). The same as above, we apply the simulator to a subsequent testing data and simulate the trend of FOs and TDIs. Then we calculate the Pearson correlation coefficient between the simulation trend line and the real. As shown in Table~\ref{tab:pcc}, the simulation trend lines of two indicators by DEMER and MAIL achieve high correlations to the real with Pearson correlation coefficient of 0.8 approximately. While the methods SUP and GAIL, trained directly with real data, get bad performance in this evaluation.

\begin{figure*}[t]
	\centering
	\includegraphics[width=0.8\textwidth]{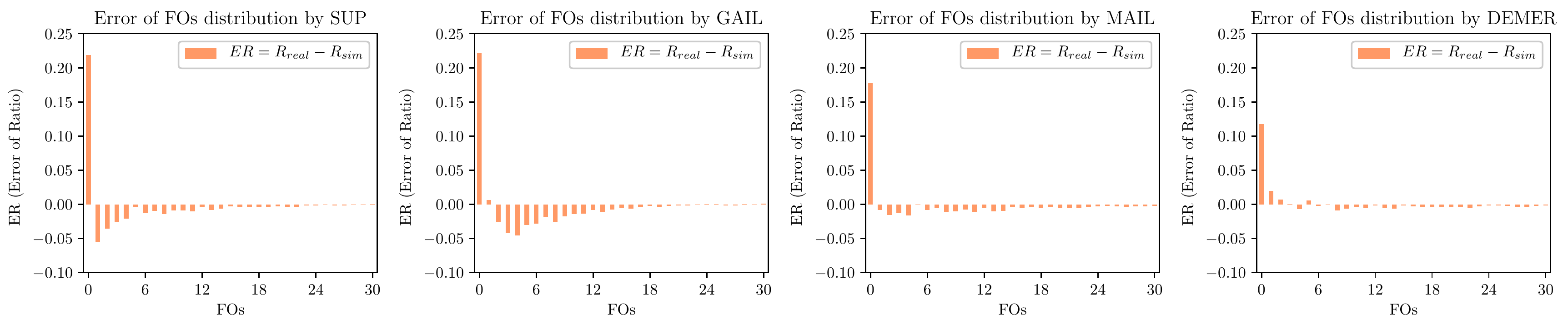}
	\caption{Error of FOs distribution generated by four different methods on testing data. Y-axis is the error of FOs distribution between the simulation and the real data. The original distribution is presented in the supplement material.}
	\label{fig:fos_dist_err}
\end{figure*}

\begin{table}[t]
	\caption{Comparison of test log-likelihood on real data.}
	\label{tab:logll}
	\begin{tabular}{ c|c|c }
		\hline
		\textbf{Methods} & \textbf{Training set} & \textbf{Testing set}\\
		\hline
		SUP & 17.09 & 18.00\\
		GAIL & 18.43 & 17.85\\
		MAIL & 15.27 & 14.52\\
		DEMER & \textbf{21.74} & \textbf{21.21}\\
		\hline
	\end{tabular}
\end{table}

\begin{table}[t]
	\caption{Comparison of Pearson correlation coefficients on FOs and TDIs trend lines.}
	\label{tab:pcc}
	\begin{tabular}{c|c|c}
		\hline
		\textbf{Methods} & \textbf{FOs} & \textbf{TDIs}\\
		\hline
		SUP & -0.0213 & 0.0010\\
		GAIL & 0.4987 & 0.4252\\
		MAIL & \textbf{0.8129} & 0.7861\\
		DEMER & 0.7945 & \textbf{0.8596}\\
		\hline
	\end{tabular}
\end{table}

\textbf{Distribution of program response}.
To compare the generalization performance of models, we apply the built simulators to subsequent program recommendation records. We simulate the drivers' responses by using real program records on testing data, then compare the simulation distribution of drivers' responses with the real distribution. Here we use FOs as an indicator. Figure~\ref{fig:fos_dist_err} shows the error of FOs distributions simulated in four simulators . The simulation distributions by SUP and GAIL are biased apparently when FOs is low. The reason is that these two methods use whole or partial real data directly for building simulators, which limits the generalization performance of simulators, and the lower FOs means the higher uncertainty, especially zero. Furthermore, the FOs distribution by DEMER is exactly closer to the real than by MAIL, where the confounder setting makes difference explicitly.

\begin{figure}[t]
	\centering
	\includegraphics[width=0.8\linewidth]{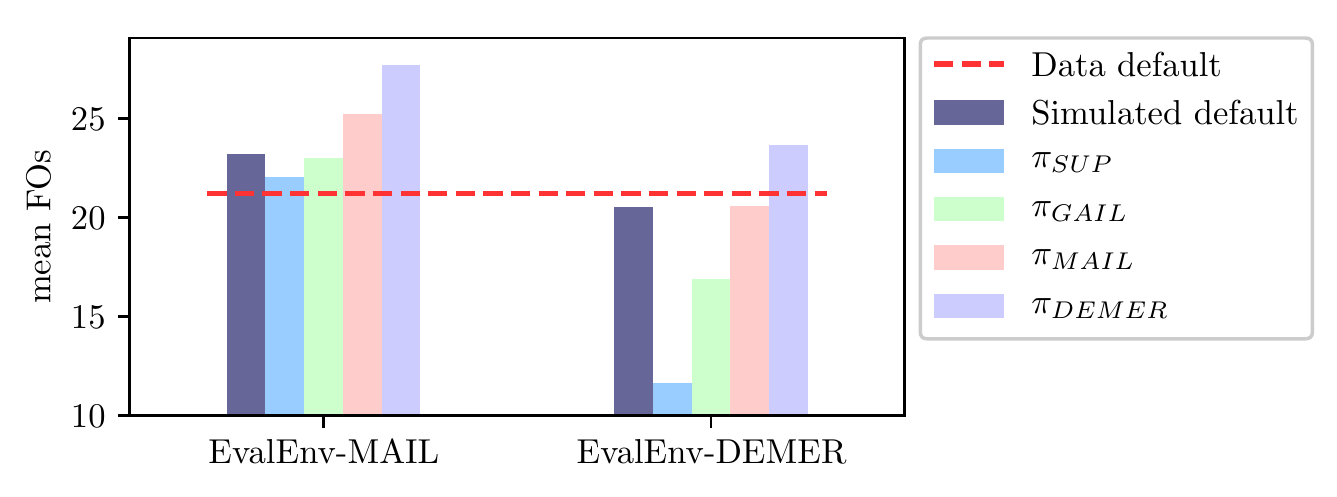} 
	\caption{Comparison of different policies trained from different simulators in \textit{EvalEnv-MAIL} and \textit{EvalEnv-DEMER}. Y-axis is the mean FOs by executing different policies. The Data default is the mean FOs in the real testing data.  The Simulated default is the mean FOs of the original simulation in each evaluation environments.}
	\label{fig:fos_rate}
\end{figure}

\textbf{Policy evaluation results in semi-online tests}.
In this part, we evaluate the effect of different simulators for reinforcement learning. Firstly, we use policy gradient method to train a recommendation policy in each simulator. Then we apply MAIL and DEMER respectively to build a virtual environment using testing data for policy evaluation, namely \textit{EvalEnv-MAIL} and \textit{EvalEnv-DEMER}. Given these two environments, we execute the optimized policies and compare the improvement of FOs. As shown in Figure~\ref{fig:fos_rate}, the policy $ \pi_{DEMER} $ optimized in the simulator built by DEMER achieves best performance on both \textit{EvalEnv-MAIL} and \textit{EvalEnv-DEMER}, while the control policies $ \pi_{SUP} $ and $ \pi_{GAIL} $ perform bad on both environments. The promotion by $ \pi_{DEMER} $ compared to $ \pi_{MAIL} $ can further verify that a virtual environment with hidden confounders can bring better performance to traditional reinforcement learning. Besides, the performance of policies $ \pi_{SUP},~\pi_{GAIL} $ shows a significant degradation  in \textit{EvalEnv-DEMER}, while not shown up in \textit{EvalEnv-MAIL}, which indicates that the environment built by DEMER can recover the real environment more precisely.

\textbf{Policy evaluation results in online A/B tests}.
We further conduct online A/B tests to evaluate the effect of the policy $\pi_{DEMER}$. The online tests are conducted in three cities of different scale. The drivers in each city are divided randomly into two groups of equal size, namely control group and treatment group. The programs for the drivers in the control group are recommended by an existing recommendation policy, which can be viewed as a baseline policy. The drivers in the treatment group are recommended by $\pi_{DEMER}$. The results of online A/B tests are shown in Table \ref{tab:abres}. The proposed policy $\pi_{DEMER}$ achieves significant improvements on FOs and TDIs in all three cities, and the overall improvements are \textbf{11.74\%} and \textbf{8.71\%} respectively.

\begin{table}[t]
	\caption{Results of online A/B tests on the platform of Didi Chuxing. Improvements of FOs and TDIs by policy \textbf{$\pi_{DEMER}$}.}
	\label{tab:abres}
	\begin{tabular}{c|c|c}
		\hline
		\textbf{Cities} & \textbf{$\Delta$FOs(\%)} & \textbf{$\Delta$TDIs(\%)}\\
		\hline
		City A & +10.73 & +6.16 \\ 
		City B & +10.16 & +9.38 \\ 
		City C & +18.47 & +17.84 \\
		\textbf{Total} & +11.74 & +8.71 \\
		\hline
	\end{tabular}
\end{table}

\section{CONCLUSION} \label{sec:ccs}
This paper explores how to construct a virtual environment with hidden confounders from observed interactions. We propose the DEMER method following the generative adversarial training framework. We design the confounder embedded policy as an important part of generator and make the discriminator compatible with two different classification tasks so as to guide the optimization of each policy precisely.
Further, we apply DEMER to build a virtual environment of driver program recommendation task on a large-scale ride-hailing platform, which is a highly dynamic and confounding environment. Experiment results verify that the policies generated by DEMER can be very similar to the real ones and have better generalization performance in various aspects. Furthermore, the simulator built by DEMER can produce better policy.
It is worth noting that the proposed method DEMER can be used not only in this task, but also in many real-world dynamic environments with hidden confounders and can lead to better learning performance.

\begin{acks}
We would like to thank Prof. Yuan Jiang for her constructive suggestions to this work.
\end{acks}

\bibliographystyle{ACM-Reference-Format}
\bibliography{confounder} 

%\clearpage
\onecolumn
\appendix
\section{Supplement Material}

\begin{figure}[h]
	\centering
	\includegraphics[width=0.6\linewidth]{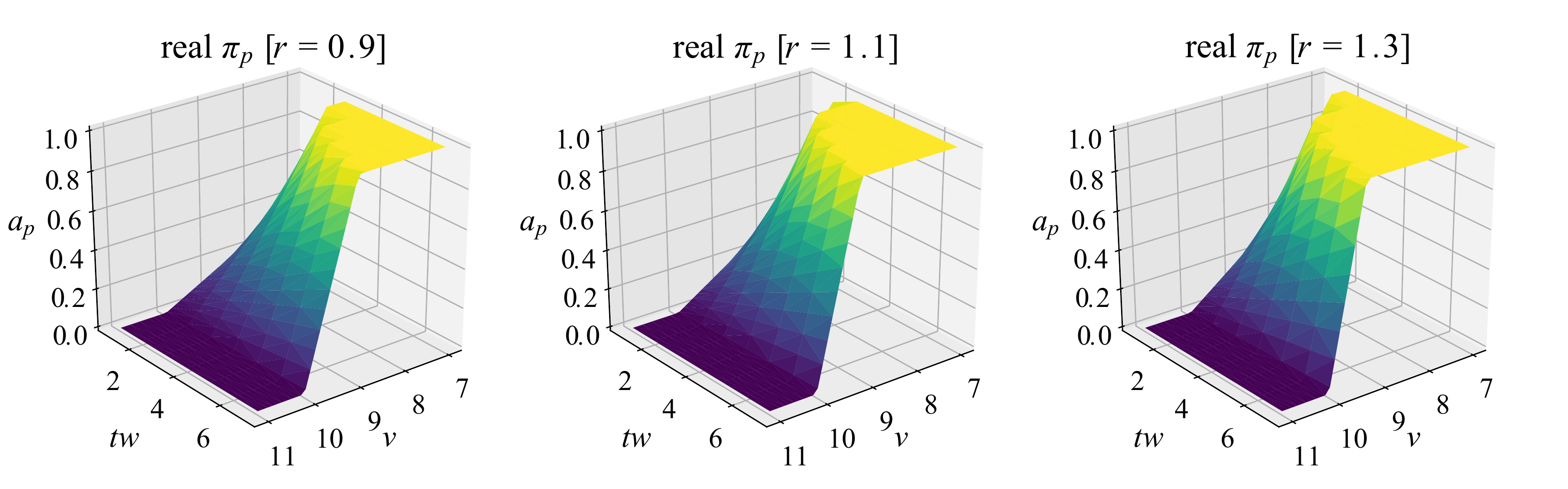}
	\includegraphics[width=0.6\linewidth]{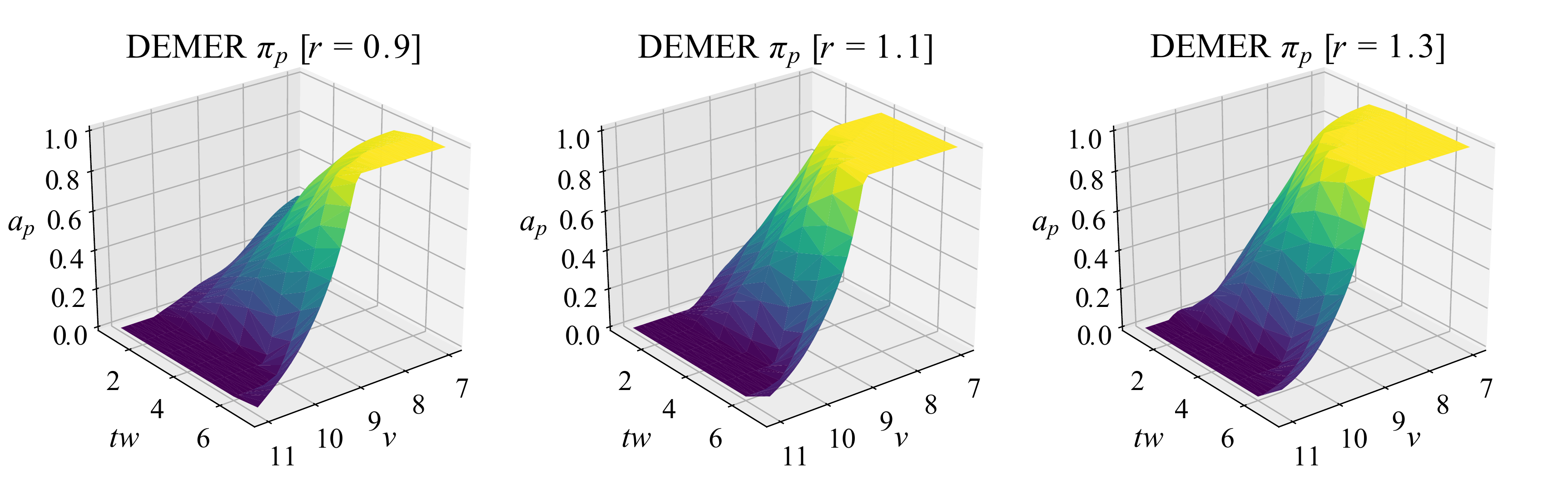}
	\includegraphics[width=0.6\linewidth]{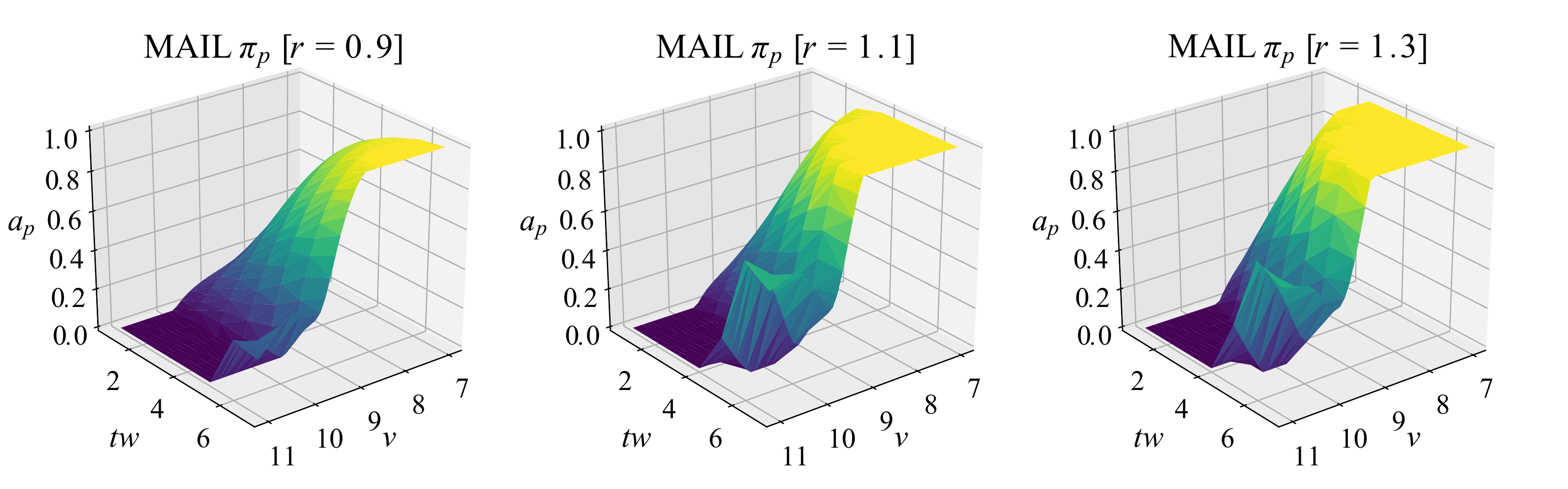}
	\caption{Visualization of the artificial platform policy function $\pi_p$ with respect to $ v $ and $ tw $ on different values of $r$. The first line is the ground-truth rule function. The second line is the policy function generated by DEMER and the third line corresponds to MAIL.}
	\label{supfig:toy_pi_p}
\end{figure}

\begin{figure}[h]
	\centering
	\includegraphics[width=0.6\linewidth]{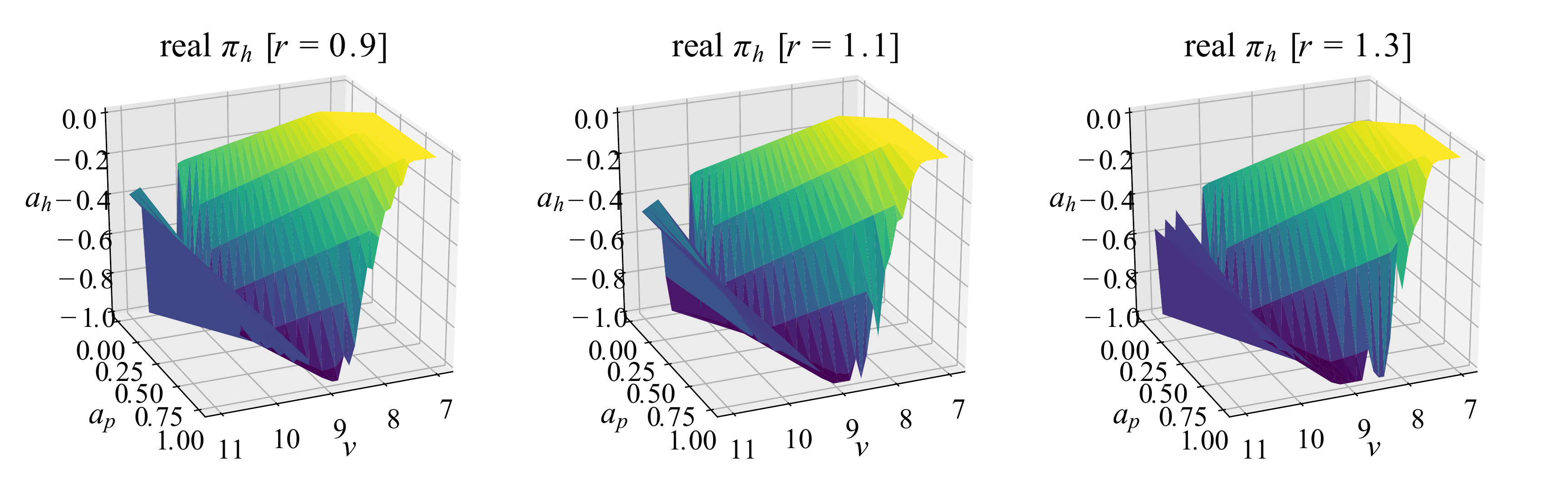}
	\includegraphics[width=0.6\linewidth]{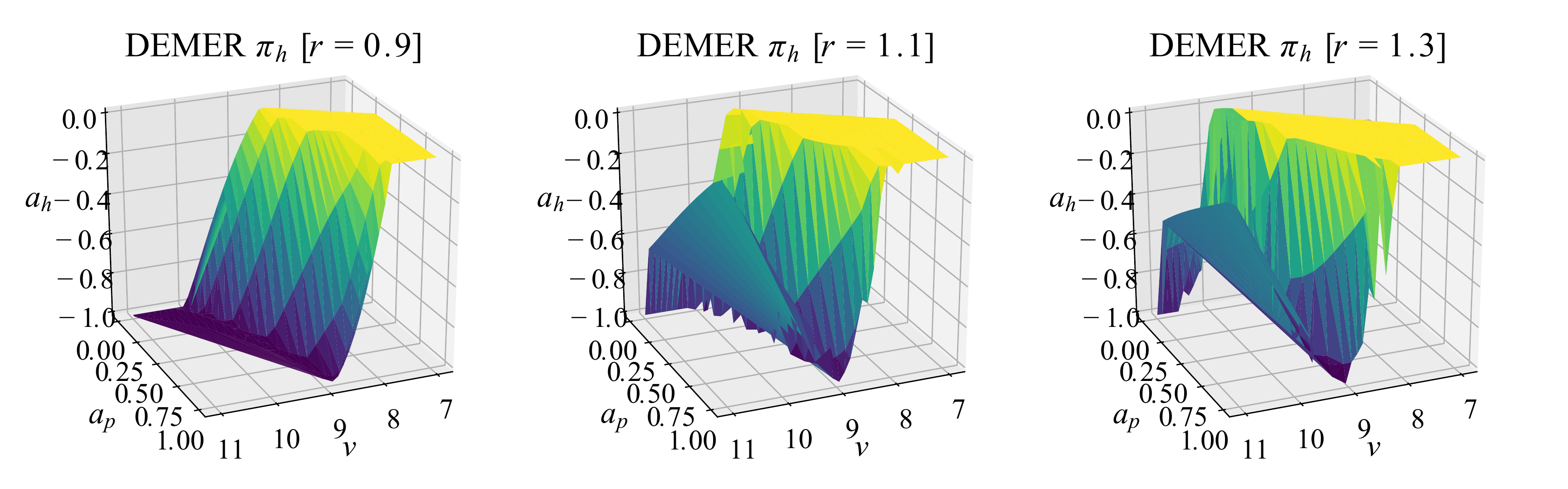}
	\caption{Visualization of the artificial confounder policy function $\pi_h$ with respect to $ v $ and $ a_p $ on different values of $r$. The first line is the ground-truth rule function. The second line is the policy function generated by DEMER.}
	\Description{The policy function map of $\pi_h$ on different value $r$. }
	\label{supfig:toy_pi_h}
\end{figure}

\begin{figure}[h]
	\centering
	\includegraphics[width=0.75\linewidth]{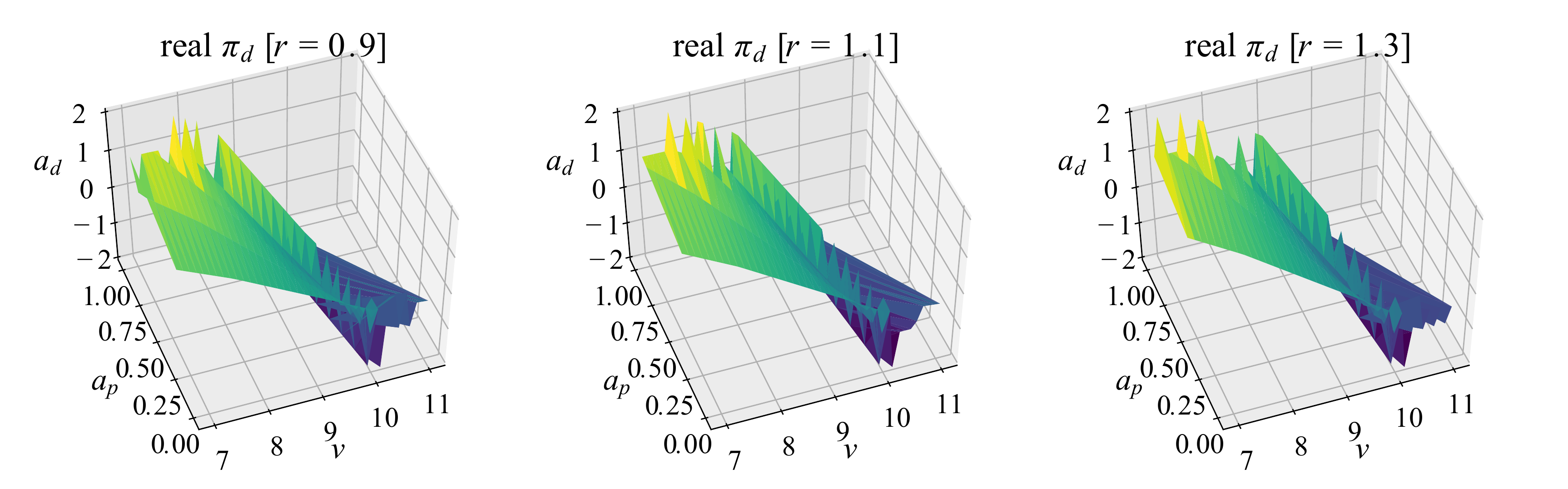}
	\includegraphics[width=0.75\linewidth]{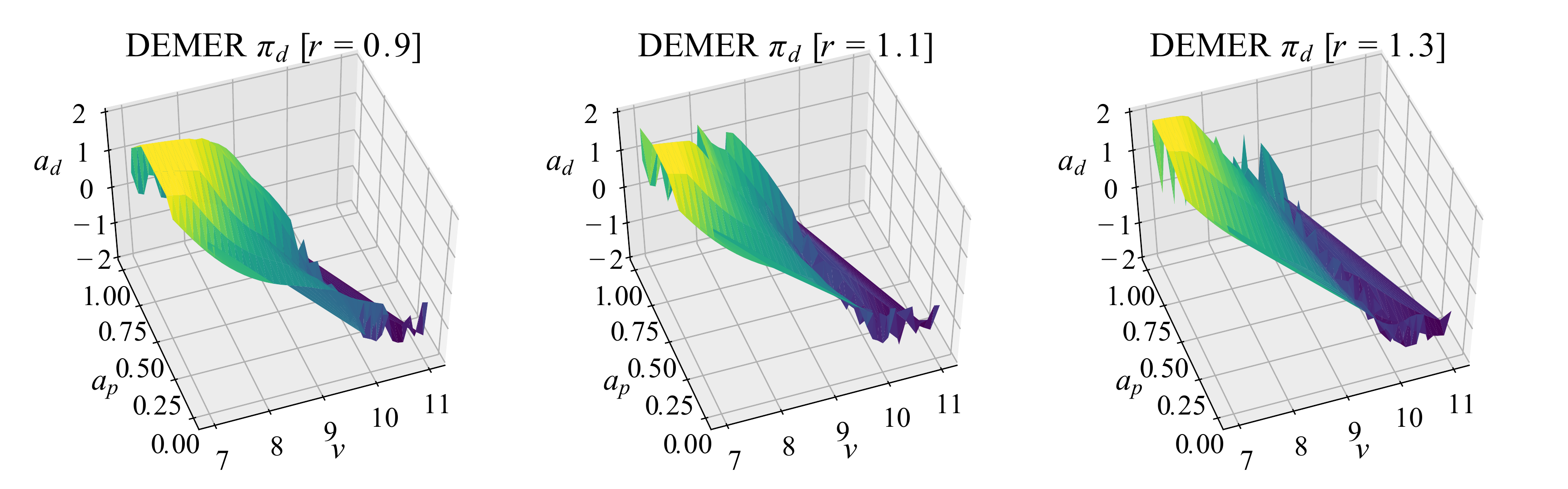}
	\includegraphics[width=0.75\linewidth]{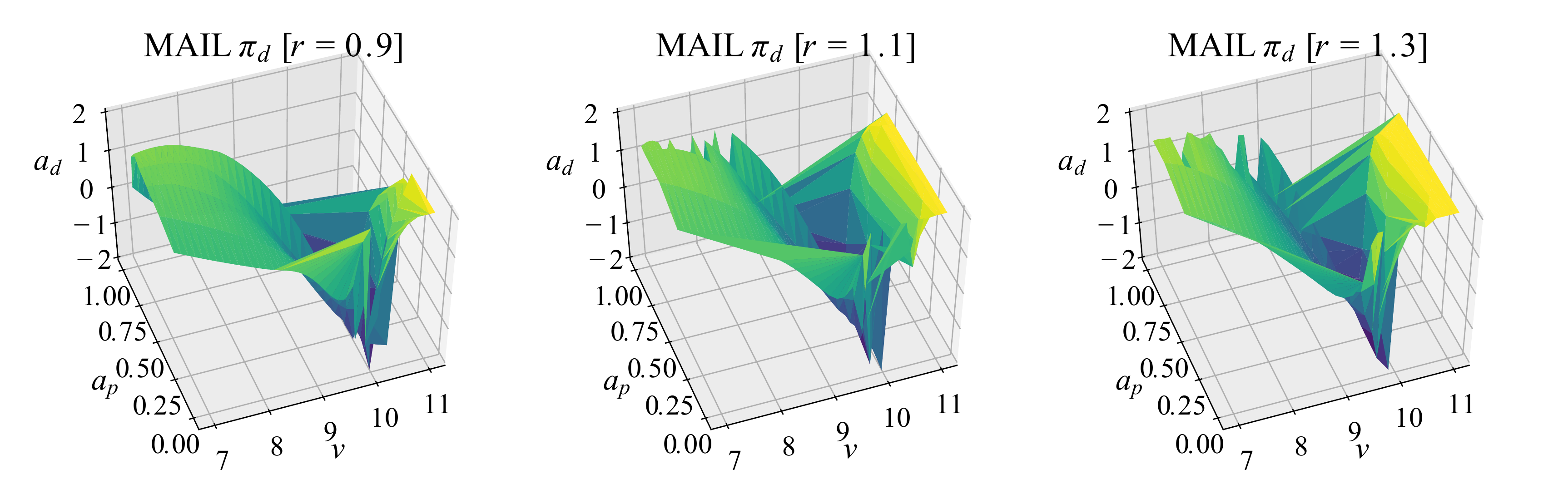}
	\caption{Visualization of the artificial driver policy function $\pi_d$ with respect to $ v $ and $ a_p $ on different values of $r$. The first line is the ground-truth rule function. The second line is the policy function generated by DEMER and the third line corresponds to MAIL.}
	\Description{The policy function map of $\pi_d$ on different value $r$. }
	\label{supfig:toy_pi_d}
\end{figure}

\begin{figure}[h]
	\centering
	\includegraphics[width=\textwidth]{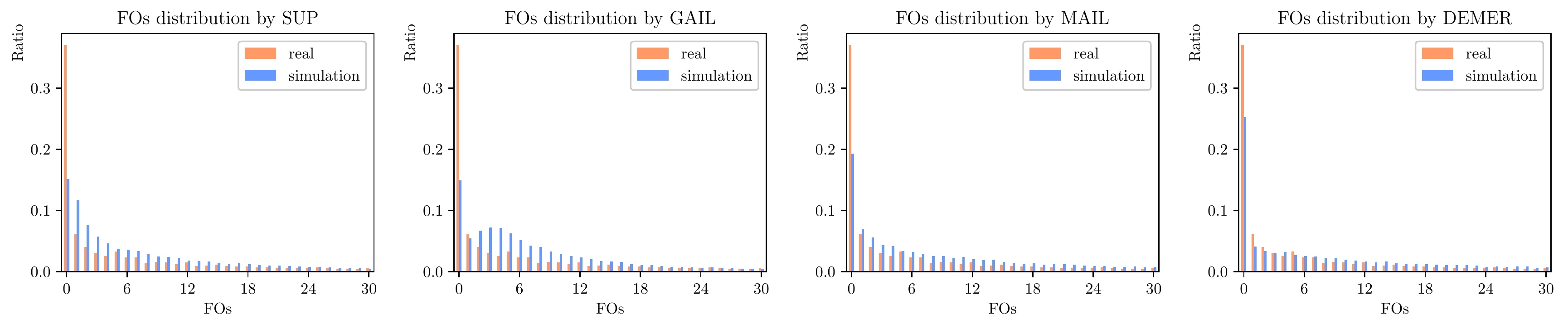}
	\caption{The original FOs distribution generated by four different methods on testing data. Y-axis is the ratio of FOs.}
	\label{supfig:fos_dist}
\end{figure}

\end{document}